\def\year{2022}\relax
\documentclass[letterpaper]{article} 
\usepackage{aaai22}  
\usepackage{times}  
\usepackage{helvet}  
\usepackage{courier}  
\usepackage[hyphens]{url}  
\usepackage{graphicx} 
\usepackage{natbib}  
\usepackage{caption} 

\usepackage{booktabs}

\DeclareCaptionStyle{ruled}{labelfont=normalfont,labelsep=colon,strut=off} 
\usepackage{algorithm}
\usepackage{algorithmic}

\usepackage{newfloat}
\usepackage{listings}
\floatstyle{ruled}
\newfloat{listing}{tb}{lst}{}
\floatname{listing}{Listing}
\title{Who Is Missing? Characterizing the Participation of Different Demographic Groups in a Korean Nationwide Daily Conversation Corpus}
\author{
    Haewoon Kwak\textsuperscript{\rm 1}, Jisun An\textsuperscript{\rm 1}, Kunwoo Park\textsuperscript{\rm 2}
}
\affiliations{
    \textsuperscript{\rm 1}Singapore Management University, Singapore\\
    \textsuperscript{\rm 2}Soongsil University, Seoul, Korea\\
    haewoon@acm.org, jisun.an@acm.org, kunwoo.park@ssu.ac.kr

}

\begin{document}

\maketitle

\begin{abstract}
A conversation corpus is essential to build interactive AI applications. 
However, the demographic information of the participants in such corpora is largely underexplored mainly due to the lack of individual data in many corpora. 
In this work, we analyze a Korean nationwide daily conversation corpus constructed by the National Institute of Korean Language (NIKL) to characterize the participation of different demographic (age and sex) groups in the corpus. 
\end{abstract}

\section{Introduction}

A conversation corpus is an essential resource for various dialogue systems, such as chatbots or virtual agents, to model real-world conversations accurately. 
There have been many attempts to collect English conversation data from various sources, such as telephone speech~\citep{williams-etal-2013-dialog}, video recording~\citep{koiso-etal-2018-construction}, social media~\citep{shang-etal-2015-neural}, 
and chat logs~\citep{lowe-etal-2015-ubuntu}. 
Non-English speaking countries, such as Japan~\citep{koiso-etal-2018-construction} and  China~\cite{shang-etal-2015-neural}, have also put effort in building  conversational corpora of their own languages.

In such conversation corpora, the participants' demographic information plays a crucial role when modeling the conversations from two perspectives: individual-level speech modeling and pairwise dialogue modeling. 
First, the speakers' background information can facilitate  generating consistent dialogue by modeling the personas of the speakers~\cite{li-etal-2016-persona}. More importantly, an imbalance of demographic groups in a dataset leads to disadvantages toward under- or overrepresented groups in algorithmic systems~\cite{koenecke2020racial,larrazabal2020gender}. 
Second, there exist demographic differences in conversations. For example, the way a teenage girl talks to her same-age friend would likely be different from how she talks to her grandmother. 
The demographic differences in pairwise conversation become more apparent in languages with honorific systems, as often used in non-English speaking countries.
An apparent example is the Korean language, which has a rich and complex honorific system, as does the Japanese language. Korean honorifics appear as ``subject honorification, object exaltation, and the six various speech levels''~\cite{wikipedia}. 
For example, the short question, ``Have you had lunch?'', could be strikingly different in character and word levels from \includegraphics[height=\fontcharht\font`\B]{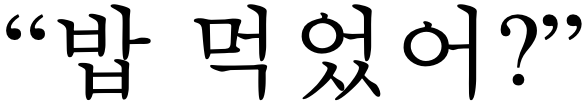} (plain speech) to  \includegraphics[height=\fontcharht\font`\B]{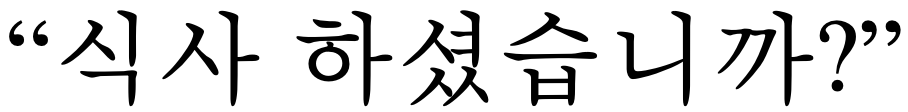} (honorific speech) or \includegraphics[height=\fontcharht\font`\B]{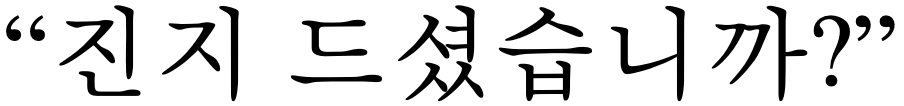} (more honorific speech). 
Thus, the participants' demographic information in a public dataset should be carefully examined if possible. Unfortunately, many existing datasets fail to provide such individual data, and  participants' demographic information in conversation corpora is underexplored.

In this work, we examine a Korean nationwide daily conversation corpus, ``Daily Conversation Corpus 2020 (DCC2020 v1.0, \includegraphics[height=\fontcharht\font`\B]{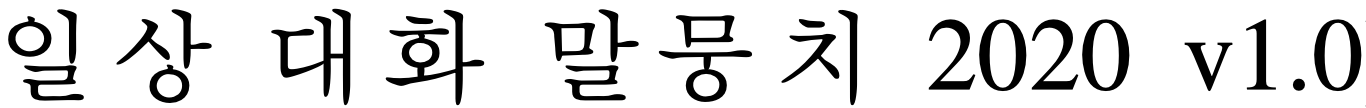},'' constructed by the National Institute of Korean Language (NIKL)~\citeyearpar{corpus_for_all}. 
The participants' demographic information is recorded with their explicit consent based on their Korean social security number, which encodes their sex and birth year. Thus, in contrast to other studies based on self-reporting or inference, we ensure a perfect accuracy of the demographic information in our study. 
We aim to understand the participation of different demographic groups and their impact on the conversations in this corpus. 

In Korea, the use of honorific language largely depends on age~\cite{yoon2004not}. Thus, we focus on (1) how balanced (or unbalanced) conversations can be collected from pairs made up of different demographic groups, and (2) how other characteristics besides honorifics (e.g., topic choice, participation time, vocabulary choice) can be affected by the inclusion of participants of different ages and sexes. 

For this, we pose four research questions that cover different stages of data collection. The first two research questions are about the pre-collection stage, and the remaining two research questions are about the characteristics of the collected conversations. 
These questions guide us to investigate DCC2020 from multiple perspectives, including participant recruitment and the characteristics of the conversations:

\begin{itemize}
\setlength\itemsep{0em}
    \item RQ1: How do demographic features influence the choice of conversation partners? ($\S$3.1)
    \item RQ2: How are demographic features associated with the  participants' topic choices? ($\S$3.2)
    \item RQ3: How do demographic features affect the   participation in conversations? ($\S$3.3)
    \item RQ4: How do demographic features affect the characteristics of conversations? ($\S$3.4)
\end{itemize}

\section{Daily Conversation Corpus 2020}
\label{sec:dcc}

DCC2020 is a 500-hour-long transcribed corpus constructed by NIKL~\shortcite{nikl}, a governmental organization aimed at ``collecting national language resources and reinforcing the integrated information service.''
It contains 2,232 conversations (1,079 with strangers and 1,153 with non-strangers) and 2,739 persons (675 males and 2,064 females). Each conversation has metadata, such as age group and sex of the participants, their relationship, and  utterances. Personally identifiable information of participants is not included.  

All the conversations were recorded. 
Before the main recordings, the participants had one or two short  prerecording sessions of 3 to 5 minutes each to check the recording conditions and to help build rapport between the stranger participants.  
The recorded conversations are later transcribed with masking all the sensitive information. Each transcriber processed at most four 15-minute recordings per day to ensure quality control. 

The participants were required to select the topic of the conversation when applying for the program among 15 generic topics, including sports and family, or among 13 news articles across the business, world, local, and culture sections. 
In total, 1,818 conversations about generic topics and 414 conversations about news articles were collected.

Compared to conversations extracted from social media, which have a relatively low number of turns~\cite{ritter-etal-2010-unsupervised} or that need to be split for topic consistency~\cite{shang-etal-2015-neural}, 
DCC2020 has the advantage that a well-trained moderator was always available at the site. 
Although the moderators usually just monitored the conversations without doing anything, they could ask the  participants to return to the topic if the conversation started going off topic.

\subsection{Recruiting Strategies}
\label{subsec:recruiting}
To get a nationwide sample of the entire Korean population, NIKL did proportional sampling across the regions and uniform sampling from (age $\times$ sex) groups. The age groups are the 10s, 20s, 30s, 40s, 50s, and 60s, and sex groups are females and males. Half of the original quota was considered the minimum to be recruited because some groups are challenging to recruit.

The recruiting was done through online channels. 
According to NIKL, their preference was for two persons, family or friends, to apply together (called non-strangers).
When one person applied alone, NIKL matched that person with another applicant who is in a similar age group and who had selected the same conversation topic. Each participant receive 20,000 KRW (around \$17 USD) for taking part in the conversation and recording.

\section{Result}

\subsection{A1: Overrepresented Conversation Pairs in the Same Demographic Group}
\label{subsec:samepair}

NIKL preferred two persons to apply together for participation, as explained in $\S$\ref{subsec:recruiting}. 
This strategy is widely acceptable for conversation data collection as two \emph{non-stranger} participants can typically have a smoother conversation. 
Thus, it is crucial to understand potential biases in non-stranger pairs to obtain representative conversations ``in the wild.''

We use the random rewiring of conversation partners to measure potential biases, which is commonly used in network science~\cite{milo2002network}. It randomly picks two pairs (A-B and C-D) and swaps their partners (A-D and C-B). After a substantial number of swaps, we count the number of conversations for each demographic pair. These numbers are the result of one null model. We build 10,000 null models and compute the Z-score of each demographic pair $p_i$ as follows:
\begin{equation}
    Z_{p_i} = \frac{N^{actual}_{p_i} - avg({N^{null}_{p_i}})}{std({N^{null}_{p_i}})}
\end{equation}
where $N_{p_i}$ is the number of conversations of $p_i$. A high (low) $Z_{p_i}$ indicates that the pair $p_i$ is over-represented (under-represented) in the actual conversations compared to a ``by chance'' occurrence.

\begin{figure}[h!]
\centering
\includegraphics[width=80mm]{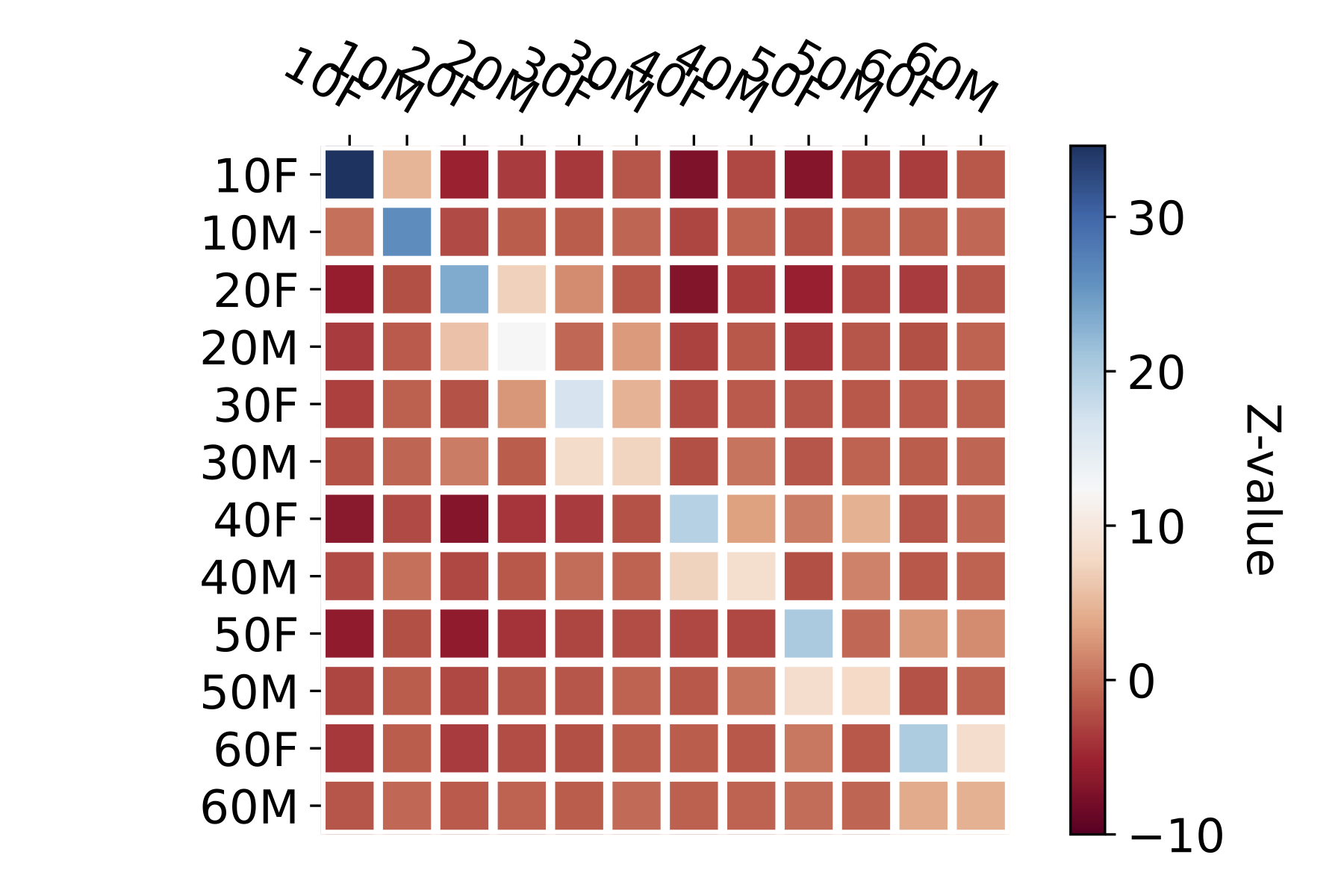}
\caption{Z-score of non-stranger pairs (label: age+gender).}
\label{fig:demographic_pair_z}
\end{figure}

Figure~\ref{fig:demographic_pair_z} shows the Z-scores of each demographic pair in the conversations between non-strangers. The larger Z-scores are located along the diagonal, showing the overrepresented conversations between non-strangers of the same demographic group. 
We note that stranger pairs are not considered in this experiment due to the artificial effect whereby NIKL matched strangers with similar ages ($\S$2.1). 
Besides the same demographic group, we identify a strong preference toward the same age group with the opposite sex, but weaker than that toward the same demographic group.

In summary, when two persons applied together to participate in dialogue collection  (non-strangers), strong preferences exist toward the same demographic group followed by the same age with the opposite sex group. This tendency should be carefully considered to obtain a balanced sample of conversations across different demographic groups. 
While NIKL matched similar age groups for the stranger pairs, someone might think that considering stranger pairs will be a solution to address the lack of conversations between certain demographic groups. We will investigate the potential of this approach by comparing the characteristics of the conversations between strangers and non-strangers in $\S$\ref{subsec:speaking_time} and $\S$\ref{subsec:vocab}.

\subsection{A2: Different Topic Preferences of the Varied Demographic Groups}

As we explained in $\S$\ref{sec:dcc}, participants chose their conversation topic in advance. So how does that  decision affect the topic diversity in DCC2020?

\begin{figure}[h!]
\includegraphics[width=80mm]{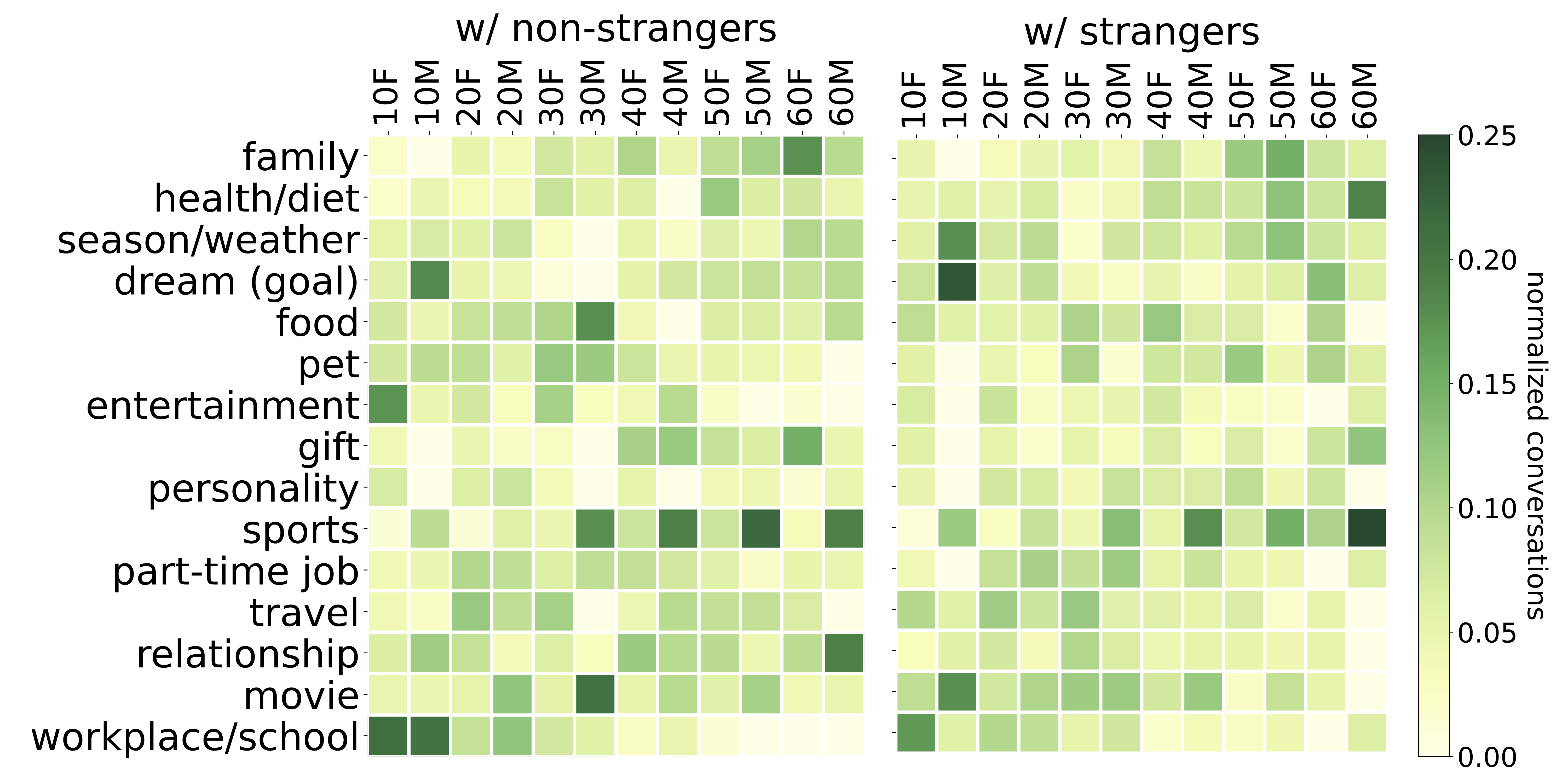}
\caption{Preferences toward generic topics} 
\label{fig:topic_pairs_ns_ws}
\end{figure}

Figure~\ref{fig:topic_pairs_ns_ws} shows the normalized topic preferences for each demographic group for when people talk to non-strangers (left) or strangers (right). 
In other words, the sum of the normalized preferences for each group across all 15 topics is 1.  
We observe some noticeable trends. For example, males express a  preference toward sports, as expected. Preferences toward family and health topics become strong as age increases. Teenage males have a strong preference toward their life dreams, but teenage females tend to prefer to talk about their school (workplace/school). 
However, the visible differences between the two heatmaps are not statistically significant for all the  demographic groups except for the 40s females, according to the chi-square test (p=0.003 for 40s females, p $>$ 0.05 for the other demographic groups).

\begin{figure}[h!]
\centering
\includegraphics[width=65mm]{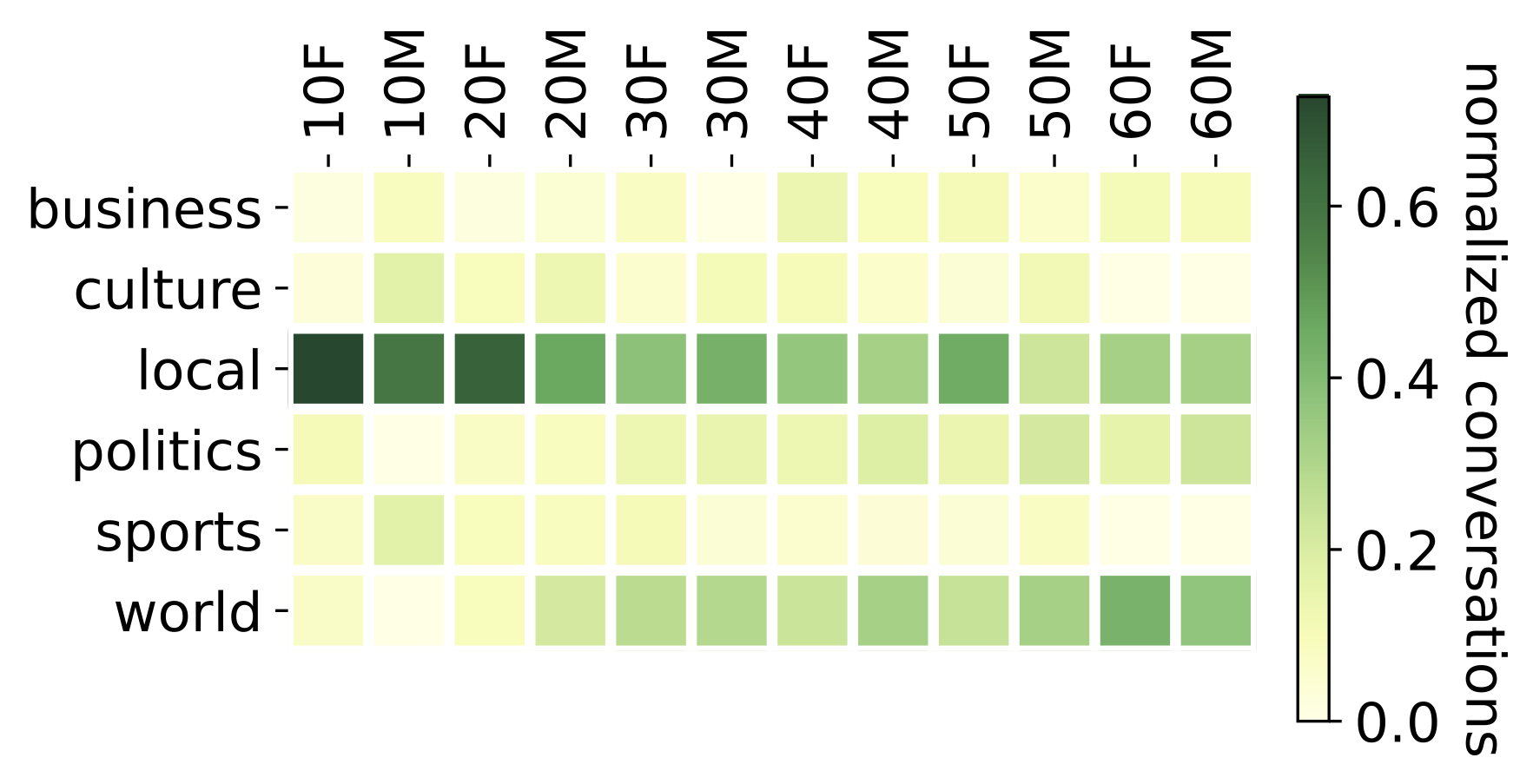}
\caption{Preferences toward the news sections}
\label{fig:newstopic_pairs}
\end{figure}

Figure~\ref{fig:newstopic_pairs} shows the normalized preferences toward the news section. As the number of participants who selected a new article as a conversation topic is much lower than those who chose a generic topic (414 vs. 1,818 conversations), 
we do not divide the result by stranger vs. non-stranger settings to avoid an overemphasis of their differences. 

By comparing with the news consumption patterns of Koreans  reported in \citet{10.1007/978-3-319-67217-5_9}, we can identify some commonalities and differences. 
In DCC2020, local news as a conversation topic is likely to be chosen by many different demographic groups, while world news is selected by relatively older generations, which are well aligned with the known news consumption patterns.  
However, a strong preference of 50s males toward politics that is observed in news consumption does not appear in the conversations in DCC2020. 
Also, females' strong preference towards cultural news in news consumption does not appear in the conversation corpus. Examining this result manually, we find that NIKL selected a news article entitled ``immigration issues are not easy to resolve but can strengthen our culture,'' which might not be considered typical culture news, such as books, music, arts, or films~\cite{guardian_culture}. 

In summary, different demographic groups show different topic preferences for the conversations. 
Thus, in combination with the over-represented conversation pairs with the same demographic, this  suggests the topic distribution in a conversation dataset can be biased. Participants' choice of conversation topics can further aggravate this  bias.

\subsection{A3: Decreasing Participation of the Varied Demographic Groups in Conversations with Strangers}
\label{subsec:speaking_time}

In $\S$\ref{subsec:samepair}, we find that, in a non-stranger setting, a conversation pair is more likely to be formed from the same demographic group, which hinders making balanced conversational data. Then, can `matching strangers of different demographic groups' be a solution to  this issue? 

\begin{figure}[h!]
\includegraphics[width=80mm]{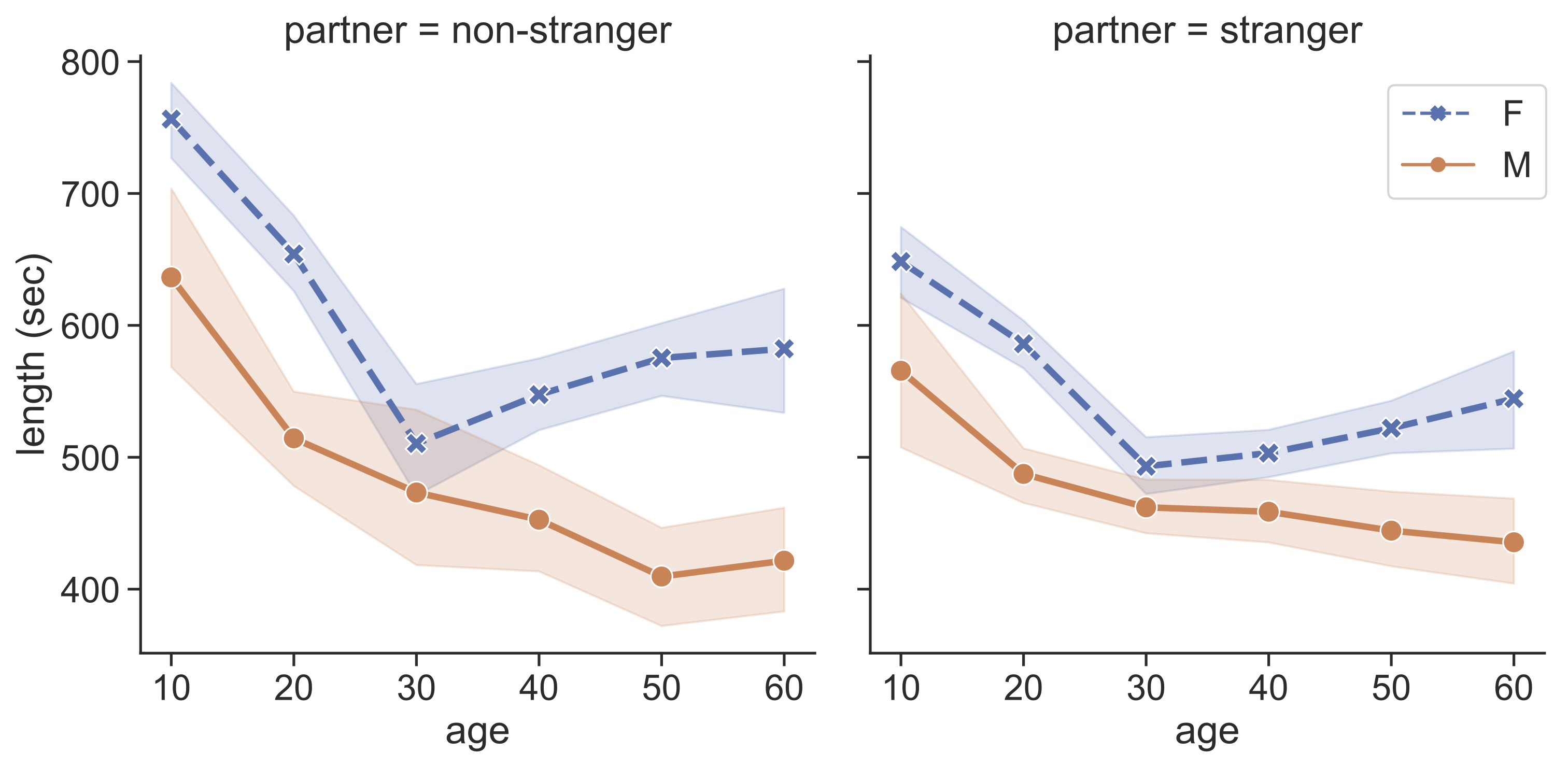}
\caption{Speaking time of each demographic group} 
\label{fig:speaking_time}
\end{figure}

Figure~\ref{fig:speaking_time} shows the speaking time of a participant of a specific demographic group when they talk to strangers and non-strangers. We compute the speaking time using the start and end timestamps of the utterances of each participant. We find a sex difference in adapting to conversations with strangers by one-sided t-test. Females of all age groups except for the 30s and 60s show significantly decreased speaking time when talking to strangers (p $<$ 0.005). By contrast, changes in the  males times are not significant (p $>$ 0.05).

\begin{figure}[h!]
\centering
\includegraphics[width=\columnwidth]{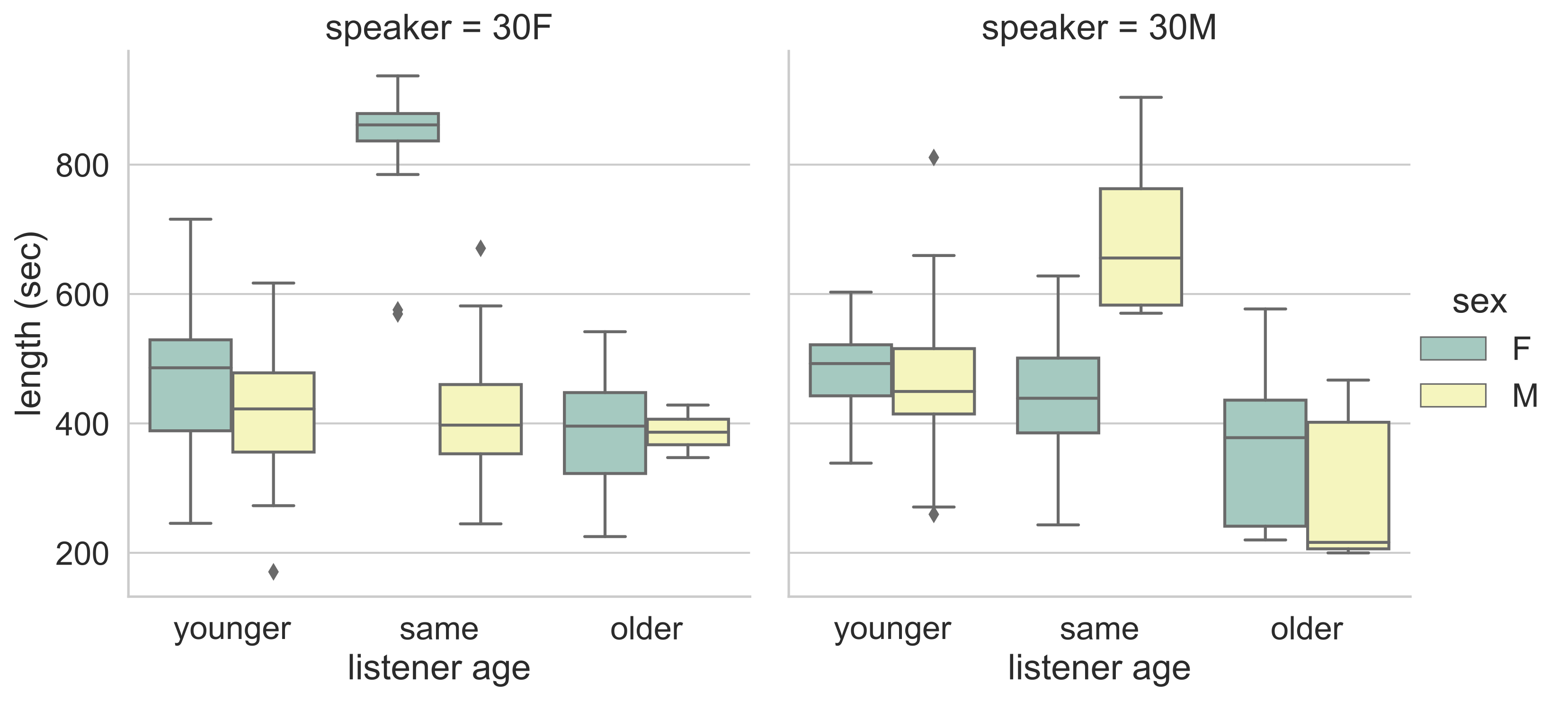}
\caption{Speaking time of the 30s group according to the listeners  (when the listeners are strangers)}
\label{fig:agetype_sp_30_length_by_sp_lt_ws}
\end{figure}

Another important factor that might influence the conversation,  particularly in Korea, is age\textemdash ``age is the first consideration when Koreans communicate with one another~\cite{doi:10.1080/01463379209369857}''. 
Figure~\ref{fig:agetype_sp_30_length_by_sp_lt_ws} shows how the 30s group adapt to conversations with strangers according to the age of the strangers. We focus on the stranger pairs to understand the effect  clearly, and the 30s were selected because they are in the middle of the age groups. 
Clearly, they demonstrate strong participation in a conversation  when they talk to the \emph{same} demographic group. This tendency is observed across all  age and sex groups. 
Moreover, the 20s and 50s females and 30s males show statistically significantly longer speaking time when they talk to younger participants than older participants, as confirmed by one-sided t-test (p $<$ 0.01), implying that the age of strangers differently affects speakers. 

The interactivity, another dimension of participation, can be measured by the number of turns. 
We compare the number of turns in conversations between strangers and that between non-strangers. 
We find that conversations between non-strangers are more interactive than conversations between strangers (p=0.012 by t-test), 
and conversations between the same age group are more interactive than different age groups (p$<$0.0001 by t-test). 

In summary, as conversations between non-strangers have a strong bias toward the same demographic or age group, conversations between strangers of different demographic groups are essential to get a balanced sample of conversations. However, strangers of different demographic groups might lead to less active, less interactive conversations. 

\subsection{A4: Increased Vocabulary  When Talking to the Same Demographic Group}
\label{subsec:vocab}

In $\S$\ref{subsec:speaking_time}, we find conversations between the same demographic group are more active than those between different demographic groups. But, how does this affect their vocabularies used in the conversations?


\begin{figure}[h!]
\centering
\includegraphics[width=\columnwidth]{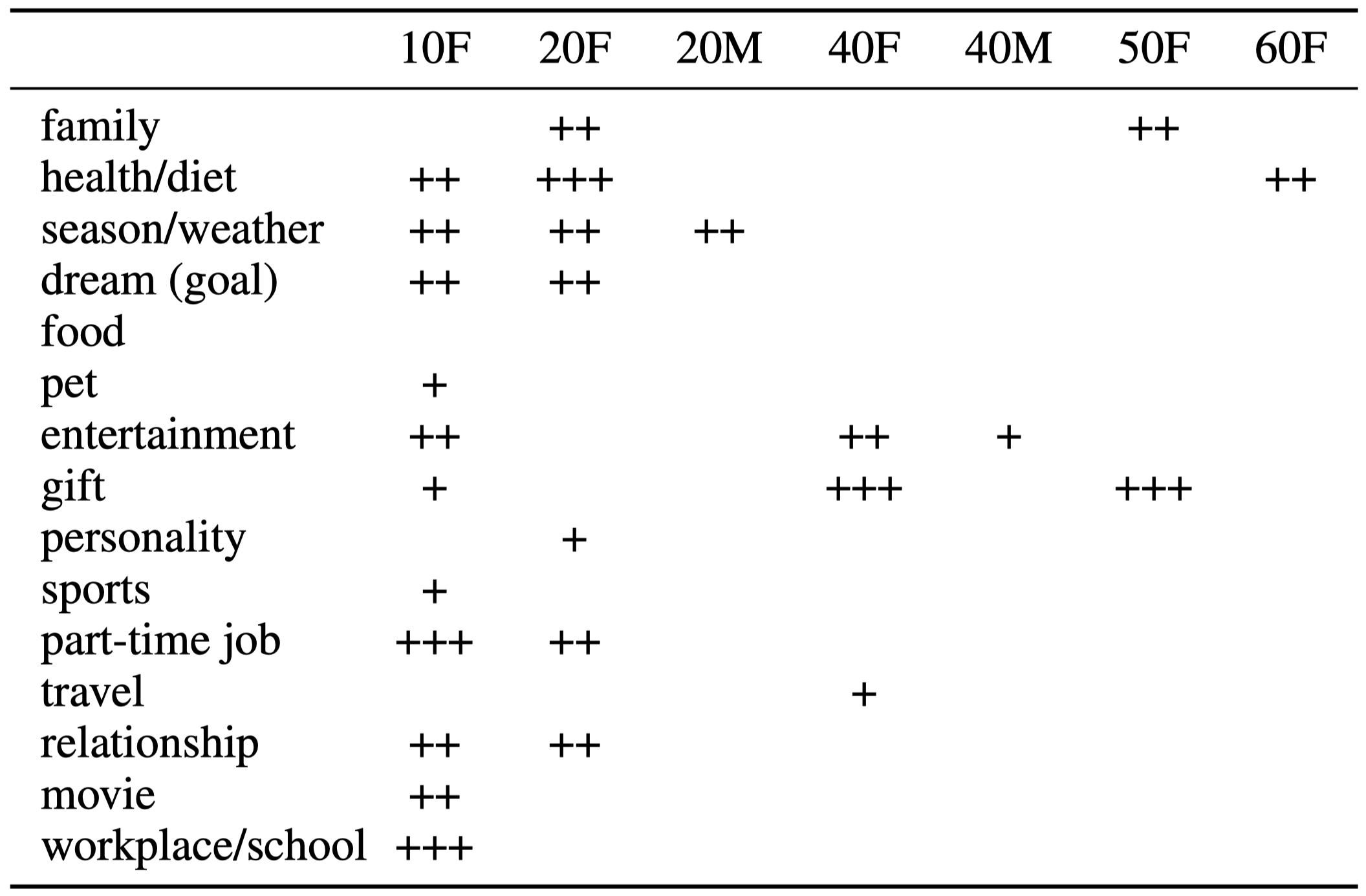}
\caption{Differences in the number of unique tokens when one demographic group (column) talks to the same group ($U_{s}$) vs. different groups ($U_{d}$).  +++: p $<$ 0.01, ++: p $<$ 0.05, +: p $<$ 0.1 when $avg(U_{s}) > avg(U_{d})$ by one-sided t-test. There are no cases of $avg(U_{s}) < avg(U_{d})$.}
\label{tbl:utoken_num_diff}
\end{figure}

Figure~\ref{tbl:utoken_num_diff} shows the result of the comparisons between the number of unique tokens when people in one demographic group (column) talk to the same demographic group versus when they talk to other demographic groups. 
We omit the demographic groups that do not have statistically significant results due to the lack of space. 
Similar to observations in $\S$\ref{subsec:speaking_time}, we find that each demographic group adapts to conversations with different demographic groups in a different way. 
Some groups (e.g., 10s males and females) actively adapt,  but others (e.g., 30s) do not show statistically significant changes. 
Also, if some demographic group adapts to conversations with different demographic groups, they always use more unique tokens when they talk to the same demographic group. No opposite effect is observed.

In summary, some demographic groups use more unique tokens in conversations with the same demographic group than those with different demographic groups. In addition to $\S$\ref{subsec:speaking_time}, one should consider that strangers of different demographic groups might lead to less diverse conversations.

\section{Discussion and Conclusion}

We examined a nationwide conversation corpus built by a governmental organization in Korea. We discovered participants' preferences toward specific demographic pairs and topics, and the impacts of these on the conversations.

Many languages use honorifics. Notably, Asian countries share the cultural context regarding social relationships influenced by age. Considering that some of those countries currently lack a large-scale corpus of their own language, there is a potential that government-related agencies may take the lead to build a nationwide corpus (due to their relatively small market). We believe that our work will be a helpful resource to guide such agencies and countries in  building a balanced nationwide corpus. 

There are some limitations in this work. First, our findings should be carefully interpreted and applied to other countries. Even though we expect that other East Asian countries might show similar trends, such as a strong preference toward the same demographic group or a longer speaking time when speaking to a younger generation, more studies are required to validate the generalizability of our findings beyond South Korea. 
Second, this work is an observational study. 
Some interventions could be found as effective that would have elicited more active, lively, and rich conversation even from stranger pairs who did not actively participate in DCC2020. 
We expect this study will become a stepping stone for future efforts toward understanding biases in conversational corpora and building fair dialogue systems.

\section*{Acknowledgments}
This research was supported by the Singapore Ministry of Education (MOE) Academic Research Fund (AcRF) Tier 1 grant. K Park was supported by the Basic Science Research Program through the National Research Foundation of Korea (NRF) funded by the Ministry of Science and ICT (No. NRF-2021R1F1A1062691).

\bibliography{main_1108}

\end{document}